\documentclass[conference]{IEEEtran}
\IEEEoverridecommandlockouts

\usepackage{cite}
\usepackage{amsmath,amssymb,amsfonts}
\usepackage{algorithmic}
\usepackage{graphicx}
\usepackage{textcomp}
\usepackage{xcolor}
\usepackage{hyperref}
\usepackage{url}
\usepackage{booktabs}
\usepackage{multirow}

\def\BibTeX{{\rm B\kern-.05em{\sc i\kern-.025em b}\kern-.08em
    T\kern-.1667em\lower.7ex\hbox{E}\kern-.125emX}}
\usepackage{etoolbox}
\newcommand{\etal}{\textit{et al.}}

\begin{document}

\title{Video-STLayout Pre-training\\
}

\author{\IEEEauthorblockN{Akash Abdu Jyothi}
\IEEEauthorblockA{\textit{School of Computing Science} \\
\textit{Simon Fraser University}\\
Burnaby, Canada \\
aabdujyo@sfu.ca}
\and
\IEEEauthorblockN{Greg Mori}
\IEEEauthorblockA{
\textit{School of Computing Science} \\
\textit{Simon Fraser University}\\
Burnaby, Canada \\
mori@cs.sfu.ca}}

\maketitle
\begin{abstract}
In recent years, pre-training has become fundamental to learning effective video representations, enabling strong transfer to downstream tasks. A popular framework in pre-training involves aligning features of a video encoder with that of another modality, for example, language or audio. We introduce Video-STLayout pre-training, a novel strategy for obtaining rich video representations informed by spatio-temporal layout of object bounding boxes. Object layouts can easily be obtained by applying an off-the-shelf object detector on the video frames. Our method uses a contrastive loss to align video features with the layout features from a trained layout encoder. We show the effectiveness of our approach in the task of activity recognition in complex scenes.
\end{abstract}   


\section{Introduction}
\label{sec:intro}

Video representation learning through pre-training is an active research area~\cite{schiappa2023self,ruan2022survey,chen2023vlp}. The paradigm of pre-training, often on a large dataset, and fine-tuning on the smaller task specific dataset has become the standard approach in many video related tasks. Pre-training techniques for videos aim to learn rich representations that capture appearance, motion patterns and spatio-temporal relationships that can help downstream tasks. 

Pre-training approaches largely fall into two categories. First, we have self-supervised representation learning for videos~\cite{qian2021spatiotemporal,benaim2020speednet,tong2022videomae,feichtenhofer2022masked,wang2022bevt,wang2023videomae} where surrogate tasks on videos are used for training. They are typically helpful for downstream tasks like video action classification, spatial or temporal action detection and action anticipation. Multi-modal pre-training for video representation learning~\cite{sun2019videobert,zhu2020actbert,ging2020coot,li2022align,wang2022object,akbari2021vatt,cheng2024videollama,tang2025video} form the second group. They use surrogate tasks that also involve other modalities like text and audio. A typical surrogate task is to to align the video features with the text or audio features of a matching description using a contrastive loss. These methods are quite beneficial for downstream tasks like video captioning, video question answering, video summarization and text-to-video retrieval.

\begin{figure}[t]
\centering
\includegraphics[width=\columnwidth]{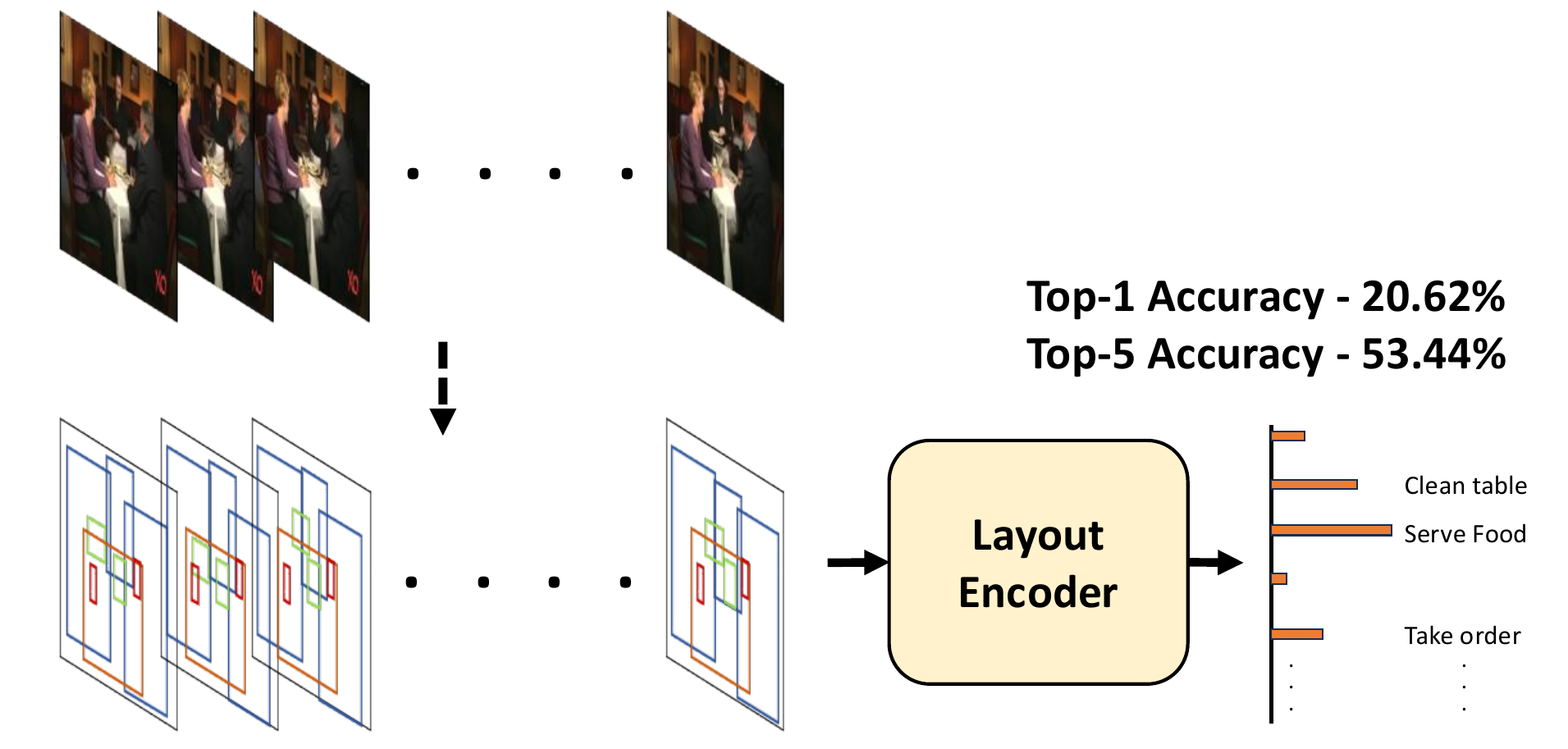}
\caption{\textbf{Unmodeled Spatiotemporal Structure in Video Encoders.} Complex activity scenes involve multiple objects whose interactions unfold over extended time horizons, often spanning tens of seconds. The figure presents validation accuracy for MOMA-LRG~\cite{luo2022momalrg} sub-activity classification using a model that relies solely on spatiotemporal bounding box layouts, without access to appearance information. The result demonstrates that object layout alone encodes rich information, motivating \textbf{layout-based supervision for pre-training video encoders} for complex activity understanding tasks. Layouts can be obtained from unlabeled videos using an off-the-shelf object detector.}
\label{fig:pull_fig}
\end{figure}

We propose Video-STLayout pre-training, a novel approach which uses spatio-temporal layout of object bounding boxes for video representation learning. As off-the-shelf object detectors have become mainstream~\cite{girshick2015fast,he2017mask,redmon2016you,liu2016ssd,carion2020end}, object bounding boxes for video frames can be easily obtained. Our main observation is the following -- a spatio-temporal layout of object bounding boxes in a video captures high level visual information that provide significant pointers regarding the activity in the video to help in video understanding tasks. For example, a layout of boxes that include \textit{persons}, \textit{chairs}, a \textit{table}, \textit{wine glasses} and \textit{dinner plates} could indicate a scene in a restaurant. Furthermore, information regarding the locations of certain bounding boxes across time could point towards the nature of the activity. For example, the activities of \textit{take order}, \textit{serve food} and \textit{cleanup table} involve distinctive spatio-temporal patterns of bounding boxes. We show the effectiveness of learned video representations in the downstream task of video activity classification.


Our pre-training approach consists of two stages to effectively learn video representations. In the first stage, we train a transformer-based \textit{layout encoder} on the downstream task (only using the training data
) using the spatio-temporal layout of bounding boxes as input, and without using the video frames. This allows the layout encoder to effectively learn useful layout representations. 
In the second stage, we perform contrastive pre-training~\cite{jia2021scaling,radford2021learning,wang2022object,li2022align} to align video representations from a \textit{video encoder} and spatio-temporal object layout representations from the trained layout encoder. 
During this stage, the video encoder is trained with the contrastive loss to distinguish between matched and mismatched video–layout pairs, while the layout encoder weights are kept frozen.
This pre-training strategy can be applied either on the target dataset or scaled to larger unlabeled video collections for improved effectiveness.
This two-stage process enables efficient video representation learning for the downstream task. We show the effectiveness of our approach in MOMA-LRG~\cite{luo2022momalrg} sub-activity classification.

In summary, we make the following contributions: (1) a novel method for exploiting easily available spatio-temporal object bounding box layouts for video pre-training, (2) a contrastive pre-training strategy using a trained layout encoder to learn video representations that are useful for video understanding tasks, and (3) experiments on MOMA-LRG~\cite{luo2022momalrg} sub-activity classification to show the benefits of our approach.

\section{Related Work}
\label{sec:rw}

\subsection{Self-supervised video representation learning} 
A variety of methods have been proposed over the years to learn video representations in a self-supervised manner, often through the design of surrogate tasks. Wang~\etal~\cite{wang2020self} proposed the task of predicting pace of the video using frames sampled at different rates while Xu~\etal~\cite{xu2019self} leveraged chronological order of the video frames. A number of highly successful pre-training strategies learn useful representations by masking and predicting masked patches in the video~\cite{tong2022videomae,feichtenhofer2022masked,wang2023videomae,gupta2023maskvit}. Sun~\etal~\cite{sun2023masked} proposed a similar approach that aims to predict motion trajectory in addition to appearance. A more detailed and nuanced presentation of research in this direction can be found in ~\cite{schiappa2023self}. Our novel approach for self-supervised learning aims to inform a video encoder with the rich information contained in the spatio-temporal layout of object bounding boxes from the video. We leverage an off-the-shelf object detector to obtain bounding boxes.

\subsection{Multimodal pre-training for video representation learning}
There have been numerous efforts in learning video representations using aligned information in other modalities like text or audio~\cite{schiappa2023self,ruan2022survey,chen2023vlp,tang2025video,gan2022vision,hu2025ophclip}. 
While a wide range of modeling techniques and loss functions have become popular in this line of research, the approach of applying contrastive learning on positive and negative pairs of video-text/audio samples is pertinent to our work. 
Miech~\etal~\cite{miech2019howto100m} introduced the \textit{HowTo100M}, a video-caption dataset  significantly larger than earlier ones and sourced from narrated instructional web videos, and showed the effectiveness of video-language representations learned using the dataset. 
Miech~\etal~\cite{miech2020end} later proposed MIL-NCE loss that incorporated multiple instance learning strategy to NCE (Noise Contrastive Estimation) loss in order to learn video-language representations effectively from visually misaligned narrations.
InfoNCE~\cite{oord2018representation} is yet another widely used contrastive loss function and plays a key role in numerous multimodal representation learning strategies~\cite{lee2024srtube,cheng2023vindlu,li2022align,yang2023tempclr,yang2023learning,wang2022object,akbari2021vatt}.
We present a novel approach for video representation learning that uses contrastive learning strategy but by exploiting the spatio-temporal layout of object bounding boxes from the video instead of a different modality like text or audio.

\subsection{Activity Recognition.}
A number of methods have been designed over the years to effectively recognize actions and activities in videos~\cite{shuchang2022survey,tang2024survey}. 
Wang~\etal~\cite{wang2018temporal} proposed an approach that divides an input video into a number of segments, and selects a random snippet from each segment to be processed by a CNN, whose outputs are later combined to obtain the final prediction. 
Feichtenhofer~\etal~\cite{feichtenhofer2019slowfast} introduced SlowFast networks for video recognition that uses two parallel convolutional pathways, one to process sparse frame samples to capture spatial information and another to process dense frames to capture temporal variations.
A number of transformer based architectures have been introduced in the recent years to become state-of-the-art choice for modeling in activity recognition~\cite{arnab2021vivit,fan2021multiscale,liu2022video,tong2022videomae,wang2023videomae}.
Our pre-training approach is tuned towards activity recognition, and can be applied on any of the above and other similar models. A spatio-temporal layout of object bounding boxes provides rich information regarding the activity in the scene. We exploit that information to learn useful video representations.

\subsection{Vision Language Models (VLMs) for Video Understanding} 
The tremendous success of large language models (LLMs) has inspired the merging of vision models and LLMs to create VLMs~\cite{chen2023video,wang2024language,maaz2024video,wang2022omnivl,wang2024omnivid,tang2025video} that cater towards a variety of video understanding tasks. VLMs aim to enhance video understanding by leveraging the representational capacity of large language models and cross-modal training. While this paradigm has proven effective for language-intensive tasks, it may be insufficient for capturing the fine-grained spatio-temporal structure present in complex activity scenes. In such settings, object-centric layouts that evolve over long temporal horizons provide complementary supervision, offering explicit cues about interactions and dynamics that extend beyond language-based signals.

\section{Method}
\label{sec:method}

\begin{figure*}[t]
\centering
\includegraphics[width=\textwidth]{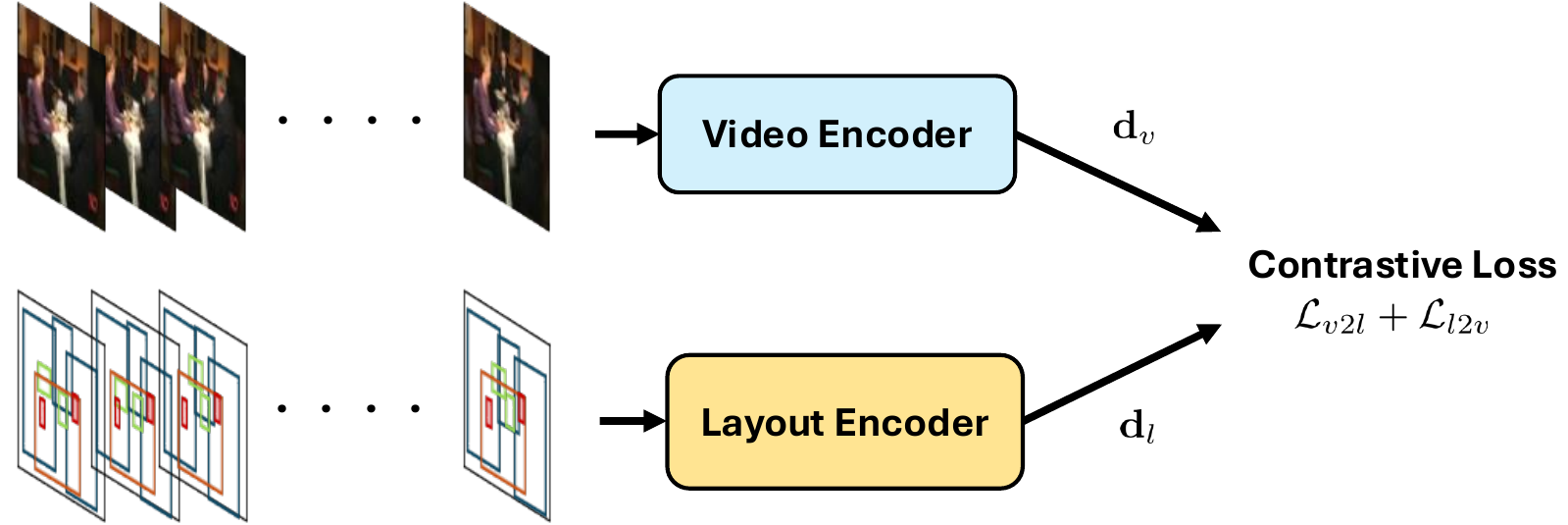}
\caption{\textbf{Video-STLayout Pre-training}. Pre-training a video encoder using spatio-temporal layouts of object bounding boxes. A trained layout encoder is used to obtain layout features. Layout encoder weights are frozen during the pre-training.}
\label{fig:pt}
\end{figure*}

In this section, we present a comprehensive overview of the Video-STLayout pre-training process. We first provide a high-level summary of the overall approach, followed by a detailed description of the video encoder architecture. Next, we outline the design and training methodology of the spatio-temporal bounding box layout encoder. Finally, we elaborate on the pre-training procedure that jointly integrates the video encoder with the trained layout encoder.

\subsection{Overview}

We are interested in pre-training a video encoder by exploiting the information from spatio-temporal layouts of object bounding boxes from the videos. 
Our approach requires a layout encoder to obtain the feature vector from a bounding box layout input. We propose a transformer based layout encoder that can capture the complex relationships between object bounding boxes across the video. As a first step, the layout encoder is trained on the downstream task dataset, using the training data, with only the bounding box layouts as input. Once the layout encoder is trained, its outputs can be compared with video encoder outputs. 

The trained layout encoder and the video encoder are then used on the downstream task dataset or a different, potentially larger, dataset of unlabeled videos for pre-training. An off-the-shelf object detector is used to obtain object bounding boxes.
In pre-training the video encoder, we apply a contrastive loss between layout features and video features, and update the video encoder weights while keeping the layout encoder weights frozen. Figure~\ref{fig:pt} depicts the pre-training process.
Video-STLayout pre-training can be applied to any video encoder model that takes a video as input and gives the target output for the downstream task. 

\subsection{Video Encoder}
We use a VideoMAE~\cite{tong2022videomae} pre-trained ViT-B~\cite{dosovitskiy2021an} as the video encoder. Given a trimmed video $v$, the video encoder takes in $N$ frames uniformly sampled across the video and outputs an unnormalized logits prediction vector $\mathbf{d}^v \in \mathbb{R}^K$, where $K$ is the number of target classes.

\subsection{Spatio-Temporal Layout Encoder}
The layout encoder takes as input the spatio-temporal bounding box layout $l$ obtained from a trimmed video $v$. $l$ consists of bounding boxes $\mathbf{b}_i, i \in \{1,...,N^b\}$, where $N^b$ is the total number of bounding boxes in $l$. 
Each bounding box $\mathbf{b}_i$ is represented as $[x_i, y_i, w_i, h_i, \mathbf{n}^f_i, \mathbf{c}^o_i]$, where $x_i$ and $y_i$ denote the top-left coordinates, and $w_i$ and $h_i$ represent the width and height. $\mathbf{n}^f_i \in \{0, 1\}^N$ is a one-hot vector indicating the frame index associated with the bounding box, and $\mathbf{c}^o_i \in \{0, 1\}^{K^o}$ is a one-hot encoding of the object class ID, where $K^o$ denotes the number of bounding box object categories.
The layout encoder is trained on the downstream task dataset, but without using the test set. This allows the encoder to learn rich representations that can be later exploited during pre-training. Similar to the video encoder, layout encoder outputs an unnormalized logits prediction vector $\mathbf{d}^l \in \mathbb{R}^K$.

We implement the layout encoder as a standard transformer. The transformer encoder takes in the set of bounding box tokens, and the decoder is executed for a single time step, using a zero-initialized target query vector of length one to aggregate the encoded representations for prediction.

\subsection{Contrastive Video-STLayout Pre-trianing }
Given a video encoder and the trained layout encoder, we can pre-train the video encoder on any unlabeled video dataset. First, we obtain spatio-temporal bounding box layouts for each video in the dataset using an off-the-shelf object detector. During the pre-training, videos are input to the video encoder and bounding box layouts are input to the layout encoder to obtain the corresponding prediction vectors. We apply a contrastive loss over these prediction vectors, and only update the weights of the video encoder while keeping the layout encoder weights frozen. 

Inspired by video-language pre-training approaches, we use $k$-pair InfoNCE loss~\cite{oord2018representation,jia2021scaling,radford2021learning,li2022align,wang2022object} for contrastive pre-training. Given a batch of $k$ pairs of video and layout prediction vectors, the contrastive loss considers the matched pair as positive sample and all other pairs as negative samples. The loss is given by the sum of the following two symmetric loss terms:

\begin{equation}
    \mathcal{L}_{v2l} = - \frac{1}{k} \sum_{i=1}^k \log \frac
            {\exp\Bigl(S({\mathbf{d}^v}_i, {\mathbf{d}^l}_i) \bigl/ \sigma\Bigl)}
            {\sum_{j=1}^k\exp\Bigl(S({\mathbf{d}^v}_i, {\mathbf{d}^l}_j) \bigl/ \sigma\Bigl)}
\end{equation}

\begin{equation}
    \mathcal{L}_{l2v} = - \frac{1}{k} \sum_{i=1}^k \log \frac
            {\exp\Bigl(S({\mathbf{d}^l}_i, {\mathbf{d}^v}_i) \bigl/ \sigma\Bigl)}
            {\sum_{j=1}^k\exp\Bigl(S({\mathbf{d}^l}_i, {\mathbf{d}^v}_j) \bigl/ \sigma\Bigl)}
\end{equation}
where $S(.,.)$ denotes cosine similarity function, and $\sigma$ is the temperature parameter that we set to $0.05$. The terms $\mathcal{L}_{v2l}$ and $\mathcal{L}_{l2v}$ represent the video-to-layout classification loss and the layout-to-video classification loss, respectively.


\section{Experiments}
\label{sec:experiments}

Through extensive experiments, we demonstrate the validity and practical relevance of Video-STLayout pre-training. We first describe the dataset used in our experiments, followed by details of the layout encoder training and Video-STLayout pre-training procedures. Finally, we present fine-tuning experiments on the downstream task, demonstrating the benefits of Video-STLayout pre-training. We implemented all experiments in PyTorch. Code is made available at \url{https://github.com/ajakash/Video-STLayout-Pretraining}.

\subsection{Dataset}
We evaluate the effectiveness of Video-STLayout pre-training on the downstream task of sub-activity video classification in the MOMA-LRG dataset~\cite{luo2022momalrg}. MOMA-LRG contains videos of complex activities often involving multiple people and objects. From the dataset’s multi-level temporal annotations, we select the trimmed sub-activity videos for our experiments. There are 91 sub-activities defined in MOMA-LRG, with a total of 15,842 samples in the dataset across all splits (9999 in train, 2657 in val and 3186 in test). Lengths of the trimmed videos range from 4s to 32s. We perform pre-training using both the training and validation videos from the same dataset.

\noindent \textbf{Video data.}
We adopt TSN~\cite{wang2018temporal} uniform sampling of 16 frames for both pre-training and fine-tuning, consistent with the Something-Something V2 configuration in the VideoMAE~\cite{tong2022videomae} codebase~\footnote{\url{https://github.com/MCG-NJU/VideoMAE/tree/main}\label{fn:videomae}} used in our experiments.
Final inference during fine-tuning on the test set is performed using 5 clips $\times$ 3 crops.

\noindent \textbf{Layout data.}
We conduct two sets of experiments: the first utilizes the annotated bounding boxes, while the second employs bounding boxes obtained from an off-the-shelf object detector. MOMA-LRG dataset contains bounding box annotations on one frame every second, with 26 actor classes and 227 object classes (total 253 bounding box classes). The number of annotated bounding boxes per sub-activity video range from 0 to 1142. For experiments using off-the-shelf object detector, we use Faster-RCNN (X101-FPN)~\cite{ren2015faster,lin2017feature,xie2017aggregated} from the detectron2 suite of models~\footnote{\url{https://github.com/facebookresearch/detectron2/blob/main/MODEL_ZOO.md}}. We apply the object detector on the same frames that were annotated. Faster-RCNN object detector covers 80 classes from the COCO dataset~\cite{lin2014microsoft}. The number of detected bounding boxes per sub-activity video range from 0 to 1187. Bounding box coordinates are normalized based on video frame height.

\begin{table}[t]
    \caption{Training configuration for layout encoder (LE) training, Video-STLayout pre-training, and fine-tuning.}
    \centering
    \small
    \setlength{\tabcolsep}{4pt} 
    \begin{tabular}{lccc}
        \toprule
        Config. & LE training & Pre-training & Fine-tuning \\
        \midrule
        Optimizer & Adam & \multicolumn{2}{c}{AdamW} \\
        Learning Rate & 1e-4 & \multicolumn{2}{c}{1e-5} \\
        Weight Decay & - & \multicolumn{2}{c}{0.05} \\
        $\beta_1$, $\beta_2$ & - & \multicolumn{2}{c}{0.9, 0.999} \\
        Batch Size & 32 & \multicolumn{2}{c}{4} \\
        Epochs & 2000 & 2000 & 150 \\
        \bottomrule
    \end{tabular}
    \label{tab:training-config}
\end{table}

\subsection{Layout Encoder Training}
We select sub-activity video samples with $\geq$10 object bounding boxes for training the layout encoder. Under this criterion, out of 9,999 sub-activity samples in the training set, we retain 9,858 samples with annotated bounding boxes and 9,766 samples with detected bounding boxes. Table~\ref{tab:training-config} summarizes the training configuration for the layout encoder, and Table~\ref{tab:LE-architecture} details the model architecture. We train the layout encoder on the training split and use validation set performance to select the checkpoint. Layout encoder was trained on 1$\times$V100 GPU for 1 day. Table~\ref{tab:LET-PT-performance} presents the validation accuracies of the layout encoder after training.

\begin{table}[t]
    \caption{\textbf{Layout encoder transformer architecture.} The transformer encoder and decoder have the same number of layers. The input dimensionality is computed as the sum of the number of object class labels, the maximum number of frames (32), and the bounding box coordinate dimension (4).}
    \centering
    \small
    \setlength{\tabcolsep}{4pt} 
    \begin{tabular}{lcc}
        \toprule
        Architecture spec. & Ann. box layout & Det. box layout \\
        \midrule
        Input dim. & 253+32+4=289 & 80+32+4=116 \\
        Output dim. & 91 & 91 \\
        Hidden dim. & 256 & 96 \\
        No. of layers & 3 & 2 \\
        No. of heads & 4 & 3 \\
        \bottomrule
    \end{tabular}
    \label{tab:LE-architecture}
\end{table}

\subsection{Video Encoder Pre-training}
We use a VideoMAE~\cite{tong2022videomae} pre-trained 12 layer ViT-B/16~\cite{dosovitskiy2021an} as the starting checkpoint for the video encoder. The pre-training and fine-tuning protocols match Something-Something V2 fine-tuning configuration from the VideoMAE codebase\footref{fn:videomae}, with the only difference being the number of epochs for pre-training. Details of the training configuration are listed in Table~\ref{tab:training-config}. 

We use the train and validation splits for pre-training, and use validation set performance to select the video encoder checkpoint for fine-tuning. Note that evaluating on the validation set does not require training additional layers, as the video encoder directly outputs unnormalized logits for the downstream task. During pre-training, we use all samples without enforcing the criterion of samples having $\geq$10 bounding boxes. Pre-training was conducted on 4$\times$V100 GPUs for 7 days, with the best-performing checkpoints typically observed after 4 days. Table~\ref{tab:LET-PT-performance} presents the validation accuracies of the video encoder after pre-training.


\begin{table}[t]
\caption{\textbf{Pre-training performance.} Validation accuracies of layout encoder (LE) and video encoder(VE) after layout encoder training and Video-STLayout pre-training, respectively. Accuracies are reported for both annotated (Ann.) and detected (Det.) bounding box layouts.}
\centering
\begin{tabular}{c c c c c}
\toprule
\multirow{2}{*}{Phase} & \multicolumn{2}{c}{Top-1 Val Acc (\%)} & \multicolumn{2}{c}{Top-5 Val Acc (\%)} \\
\cmidrule(lr){2-3} \cmidrule(lr){4-5}
& Ann. & Det. & Ann. & Det. \\
\midrule
\shortstack{LE training \\ (LE acc.)} & 20.62 & 18.99 & 53.44 & 48.19 \\
\shortstack{Pre-training \\ (VE acc.)}  & 10.26 & 7.73 & 28.52 & 12.97 \\
\bottomrule
\end{tabular}
\label{tab:LET-PT-performance}
\end{table}

\subsection{Fine-Tuning}
\begin{figure*}[t]
\centering
\includegraphics[width=\textwidth]{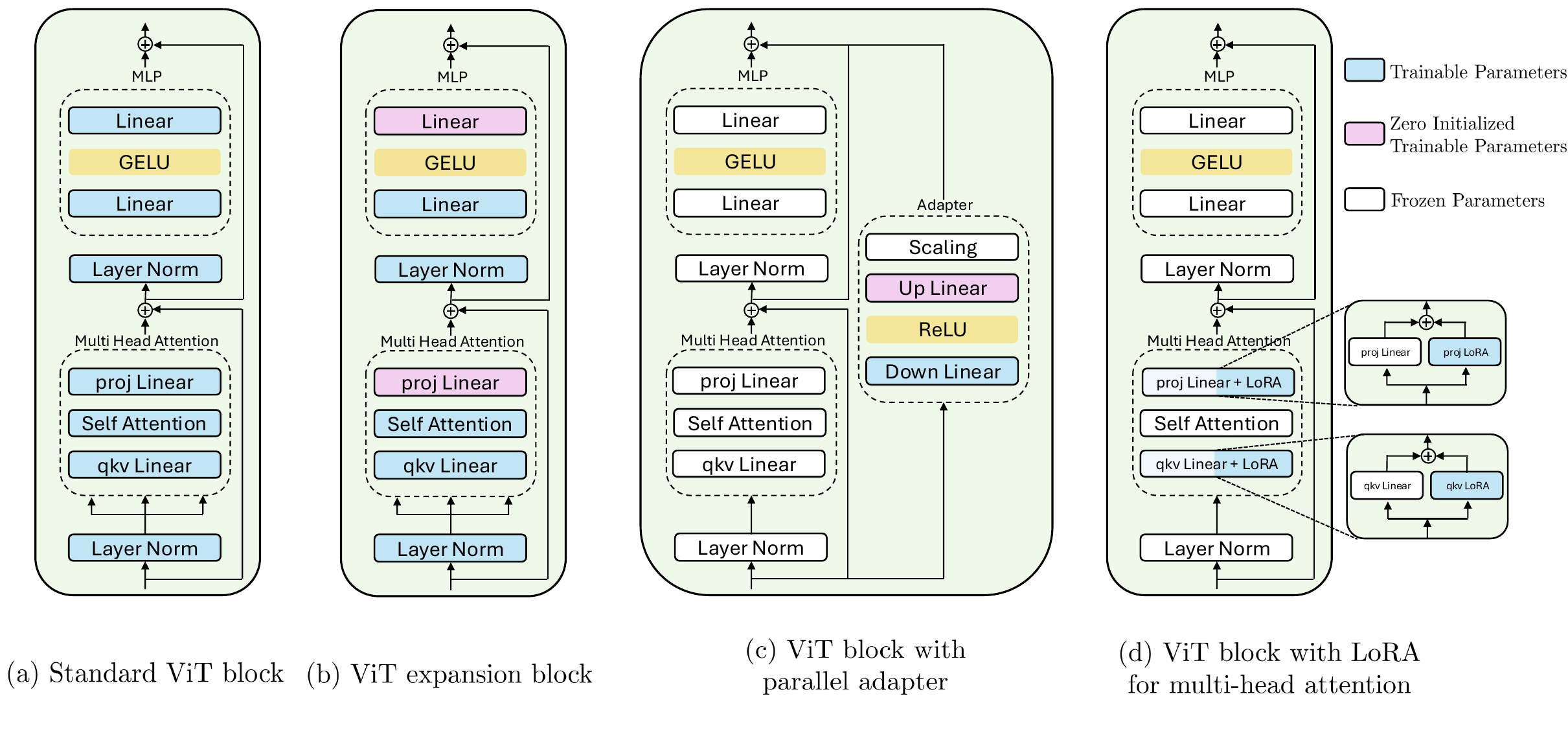}
\caption{\textbf{Parameter Efficient Fine-tuning approaches}. We employ multiple fine-tuning approaches to evaluate the effectiveness of Video-STLayout pre-training on the downstream task: (a) Standard fine-tuning where all layers of ViT blocks are trained. (b) An expanded ViT block formed by copying a existing block and appended to it sequentially. Linear layers on the output side are zeroed out to enable identity mapping. Only the expanded blocks are trained during fine-tuning, while parameters in all the original blocks are frozen. (c) ViT block with an adapter in parallel. Only adapter parameters are trained during fine-tuning. (d) ViT multi-head attention block with Low Rank Adapter (LoRA). Only LoRA parameters are trained during fine tuning. }
\label{fig:peft}
\end{figure*}

\begin{table*}[t]
    \caption{\textbf{Fine-tuning performance.} Effectiveness of Video-STLayout pre-training (V-STLayout PT) after fine-tuning (FT) on the downstream task of MOMA-LRG sub-activity classification. We perform two sets of experiments: pre-training using  (1) annotated object bounding boxes (Ann.) and (2) bounding boxes produced by an off-the-shelf detector (Det.). We conduct both full-model fine-tuning and parameter-efficient fine-tuning (PEFT).}
    \centering
    \small
    \setlength{\tabcolsep}{2pt}
    \begin{tabular}{l c c ccc ccc}
        \toprule
        \multirow{2}{*}{} &
        \multirow{2}{*}{PEFT Details} &
        \multirow{2}{*}{\shortstack{No. of trainable \\ parameters \\ (Total parameters)}} &
        \multicolumn{3}{c}{Top-1 Acc (\%)} &
        \multicolumn{3}{c}{Top-5 Acc (\%)} \\
        \cmidrule(lr){4-6} \cmidrule(lr){7-9}
        & & &
        \shortstack{V-STLayout PT \\ (Ann.) $\rightarrow$ FT} &
        \shortstack{V-STLayout PT \\ (Det.) $\rightarrow$ FT} &
        FT &
        \shortstack{V-STLayout PT \\ (Ann.) $\rightarrow$ FT} &
        \shortstack{V-STLayout PT \\ (Det.) $\rightarrow$ FT} &
        FT \\ 
        \midrule

        \shortstack{Full Model \\ Fine-tuning} & 
        -- &
        \shortstack{86,297,179 \\ (86,297,179)} &
        \textbf{50.38} & 45.48 & \underline{49.84} &
        \underline{87.82} & 86.32 & \textbf{89.52} \\[3pt]
        \midrule

        \multirow{3}{*}{Block Expansion} &
        \shortstack{expand block \\ 12} &
        \shortstack{7,087,104 \\ (93,384,283)} &
        \textbf{42.59} & \underline{36.38} & 29.44 &
        \textbf{83.77} & \underline{75.08} & 65.79 \\[3pt]

        & \shortstack{expand blocks \\ 12, 11, 10} &
        \shortstack{21,261,312 \\ (107,558,491)} &
        \textbf{47.87} & \underline{39.92} & 33.02 &
        \textbf{87.98} & \underline{83.05} & 75.74 \\[3pt]

        & \shortstack{expand blocks \\ 12, 11, 10, 9, 8} &
        \shortstack{35,435,520 \\ (121,732,699)} &
        \textbf{49.25} & \underline{42.22} & 34.84 &
        \textbf{88.04} & \underline{84.43} & 77.75 \\[3pt]
        \midrule

        \multirow{2}{*}{\shortstack{Adapter parallel \\ to ViT block}} &
        \shortstack{bottleneck dim = 32 \\ scaling = 16} &
        \shortstack{599,424 \\ (86,896,603)} &
        \textbf{45.01} & \underline{35.94} & 26.93 &
        \textbf{86.16} & \underline{77.78} & 63.65 \\[3pt]

        & \shortstack{bottleneck dim = 64 \\ scaling = 8} &
        \shortstack{1,189,632 \\ (87,486,811)} &
        \textbf{45.10} & \underline{35.97} & 27.34 &
        \textbf{86.82} & \underline{78.19} & 64.25 \\[3pt]
        \midrule

        \multirow{2}{*}{\shortstack{LoRA for \\ multi-head attention \\ (qkv \& proj)}} &
        \shortstack{proj rank = 32 \\ qkv rank = 96 \\ scaling = 16} &
        \shortstack{4,128,768 \\ (90,425,947)} &
        \textbf{49.44} & \underline{42.12} & 30.76 &
        \textbf{88.07} & \underline{83.93} & 67.92 \\[3pt]

        & \shortstack{proj rank = 64 \\ qkv rank = 192 \\ scaling = 8} &
        \shortstack{8,257,536 \\ (94,554,715)} &
        \textbf{48.93} & \underline{42.18} & 32.74 &
        \textbf{87.57} & \underline{84.37} & 68.36 \\[3pt]
        
        \bottomrule
    \end{tabular}
    \label{tab:main-results}
\end{table*}

To assess the effectiveness of Video-STLayout pre-training, we perform extensive fine-tuning experiments on the MOMA-LRG sub-activity classification dataset. In our evaluation, the pre-trained ViT-B video encoder is fine-tuned using both a standard full-parameter approach and multiple parameter-efficient fine-tuning (PEFT) configurations. Our fine-tuning protocol follows the Something-Something V2 configuration provided in the official VideoMAE codebase~\footref{fn:videomae}. The fine-tuning configuration details are outlined in Table~\ref{tab:training-config}. Fine-tuning for all variants was performed on 4$\times$V100 GPU for 12 hours.

\subsubsection*{PEFT Approaches}
\noindent \textbf{Block Expansion.} Block expansion~\cite{wu-etal-2024-llama,Bafghi_2024_CVPR} for transformers refers to the process of inserting an additional, identical transformer block in series immediately after the original one, effectively increasing network depth while preserving the pretrained block structure. Two linear layers in the newly expanded block are zero initialized to enable identity mapping. The copied blocks are fine-tuned while freezing the original blocks, thereby preserving the stability and inductive biases of the pretrained model and enabling specialized adaptation to the downstream task. Fig.~\ref{fig:peft} (b) shows an instance of an expanded block. 

\noindent \textbf{Low Rank Adaptation(LoRA).} LoRA~\cite{hu2022lora,Bafghi_2024_CVPR} applies a learnable low-rank update to a linear layer, with the output multiplied by a scaling factor. The scaling normalizes the effect of the chosen rank, ensuring that changing the rank parameter does not disproportionately alter the magnitude of the update. We apply LoRA to the two linear layers inside each multi-head attention block. Fig.~\ref{fig:peft} (d) depicts ViT-B multi-head attention blocks with LoRA.

\noindent \textbf{Adapters.} Adapters~\cite{rebuffi2017adapter,houlsby2019parameter,chen2022adaptformer} are modules that are added in series or parallel within a pre-trained model. We apply MLP adapters parallel to each ViT-B block. Each adapter consists of a down projection layer, a ReLU non-linearity at the bottleneck, and an up projection layer. Similar to LoRA, the output from the up-projection is multiplied by a scaling factor to normalize the effect of the bottleneck dimension. Fig.~\ref{fig:peft} (c) shows an instance of a ViT-B block with adapter.

\subsection{Results and Discussion}

Table~\ref{tab:main-results} summarizes the results of our fine-tuning experiments. We report both top-1 and top-5 classification accuracy for sub-activity classification on MOMA-LRG dataset. We observe that while full model fine-tuning after Video-STLayout pre-training achieves performance comparable to fine-tuning without pre-training, Video-STLayout pre-training yields consistent and significant improvements across all PEFT approaches. It is important to note that, in our experimental setup, pre-training is conducted on a relatively small dataset compared to the standard practice of large-scale pre-training. Given the gains observed under PEFT settings, we posit that scaling Video-STLayout pre-training to larger datasets could further enhance performance, potentially benefiting full model fine-tuning as well.

We further observe that pre-training with annotated object bounding box layouts yields a substantial performance boost compared to using detector-generated bounding boxes. This improvement can be attributed to the richer semantic content of the annotated layouts in MOMA-LRG, which include a significantly larger and more diverse set of object classes than those produced by Faster-RCNN detector. The increased object vocabulary results in more informative and structured layouts, enabling the video encoder to learn higher-quality representations during pre-training. In contrast, detected layouts are constrained by detector coverage and class granularity, which limits the expressiveness of the resulting representations and reduces their effectiveness for downstream adaptation.

\section{Conclusion}
\label{sec:conclusion}

In this work, we introduced a novel pre-training approach that enriched video encoder representations for complex activity understanding using spatio-temporal object layout information. We conducted experiments on the MOMA-LRG sub-activity classification task and demonstrated significant improvements in fine-tuning performance. Overall, our results highlighted the effectiveness of pre-training with the relatively underexplored modality of spatio-temporal object layouts.

\section*{Acknowledgment}
This research was enabled in part by support provided by the Digital Research Alliance of Canada (\url{https://alliancecan.ca}).

\end{document}